\documentclass[lettersize,10 pt,conference]{IEEEtran}

\IEEEoverridecommandlockouts                              

\usepackage{booktabs}       
\usepackage{amsfonts}       
\usepackage{nicefrac}       
\usepackage{microtype}      
\usepackage{xcolor}         
\usepackage{tikz,array,caption}
\usepackage{multirow}
\usepackage{float}
\usepackage{amsmath}
\usepackage{cite}
\usepackage{adjustbox} 
\usepackage{subcaption}
\usepackage{times}  
\usepackage{helvet}  
\usepackage{courier}  

\usepackage{graphicx} 

\title{\LARGE \bf
UniPLV: Towards Label-Efficient  Open-World 3D Scene Understanding by Regional Visual Language Supervision}

\author{Yuru Wang$^*$, Pei Liu$^*$, Songtao Wang, Zehan Zhang$^{\dagger}$, Xinyan Lu, Changwei Cai, Hao Li, Fu Liu, Peng Jia, Xianpeng Lang \\
\thanks{Yuru Wang and Pei Liu contributed equally to this work.}
\thanks{Yuru Wang, Songtao Wang, Zehan Zhang, Xinyan Lu, Changwei Cai, Hao Li, Fu Liu, Peng Jia, Xianpeng Lang are with Li Auto Inc., Shanghai 201800, China (e-mail: \{wangyuru, wangsongtao, zhangzehan, luxinyan, caichangwei, lihao, liufu, jiapeng, langxianpeng\}@lixiang.com).}

\thanks{Pei Liu is with The Hong Kong University of Science and Technology (Guangzhou), China (e-mail: pliu061@connect.hkust-gz.edu.cn).}
\thanks{Corresponding author: Zehan Zhang.}
}

\begin{document}
\maketitle
\thispagestyle{empty}
\pagestyle{empty}

\begin{abstract}


Open-world 3D scene understanding is a critical challenge that involves recognizing and distinguishing diverse objects and categories from 3D data, such as point clouds, without relying on manual annotations. Traditional methods struggle with this open-world task, especially due to the limitations of constructing extensive point cloud-text pairs and handling multimodal data effectively.
In response to these challenges, we present \textbf{UniPLV}, a robust framework that unifies point clouds, images, and text within a single learning paradigm for comprehensive 3D scene understanding. UniPLV leverages images as a bridge to co-embed 3D points with pre-aligned images and text in a shared feature space, eliminating the need for labor-intensive point cloud-text pair crafting. Our framework achieves precise multimodal alignment through two innovative strategies: (i) logit and feature distillation modules between images and point clouds to enhance feature coherence, and (ii) a vision-point matching module that implicitly corrects 3D semantic predictions affected by projection inaccuracies from points to pixels.
To further boost performance, we implement four task-specific losses alongside a two-stage training strategy. Extensive experiments demonstrate that UniPLV significantly surpasses state-of-the-art methods, with average improvements of 15.6\% and 14.8\% in semantic segmentation for Base-Annotated and Annotation-Free tasks, respectively. These results underscore UniPLV’s efficacy in pushing the boundaries of open-world 3D scene understanding. We will release the code to support future research and development.

\end{abstract}

\label{sec:intro}
\section{Introduction}

Open-world 3D scene understanding is a fundamental yet challenging task that aims to enable models to recognize and distinguish open-set objects and categories directly from 3D data, such as point clouds, without relying on manual annotations. Unlike traditional closed-set approaches, which depend heavily on predefined categories and human-labeled datasets, open-world methods must generalize effectively to unseen objects and categories, a capability essential for real-world applications such as autonomous driving, robotics, and augmented or virtual reality. However, the inherent lack of annotations in large-scale point cloud data and the difficulty of integrating textual semantic knowledge make open-world 3D understanding a particularly demanding goal.

\begin{figure}[t]
	\centering
	\includegraphics[width=1.0\linewidth]{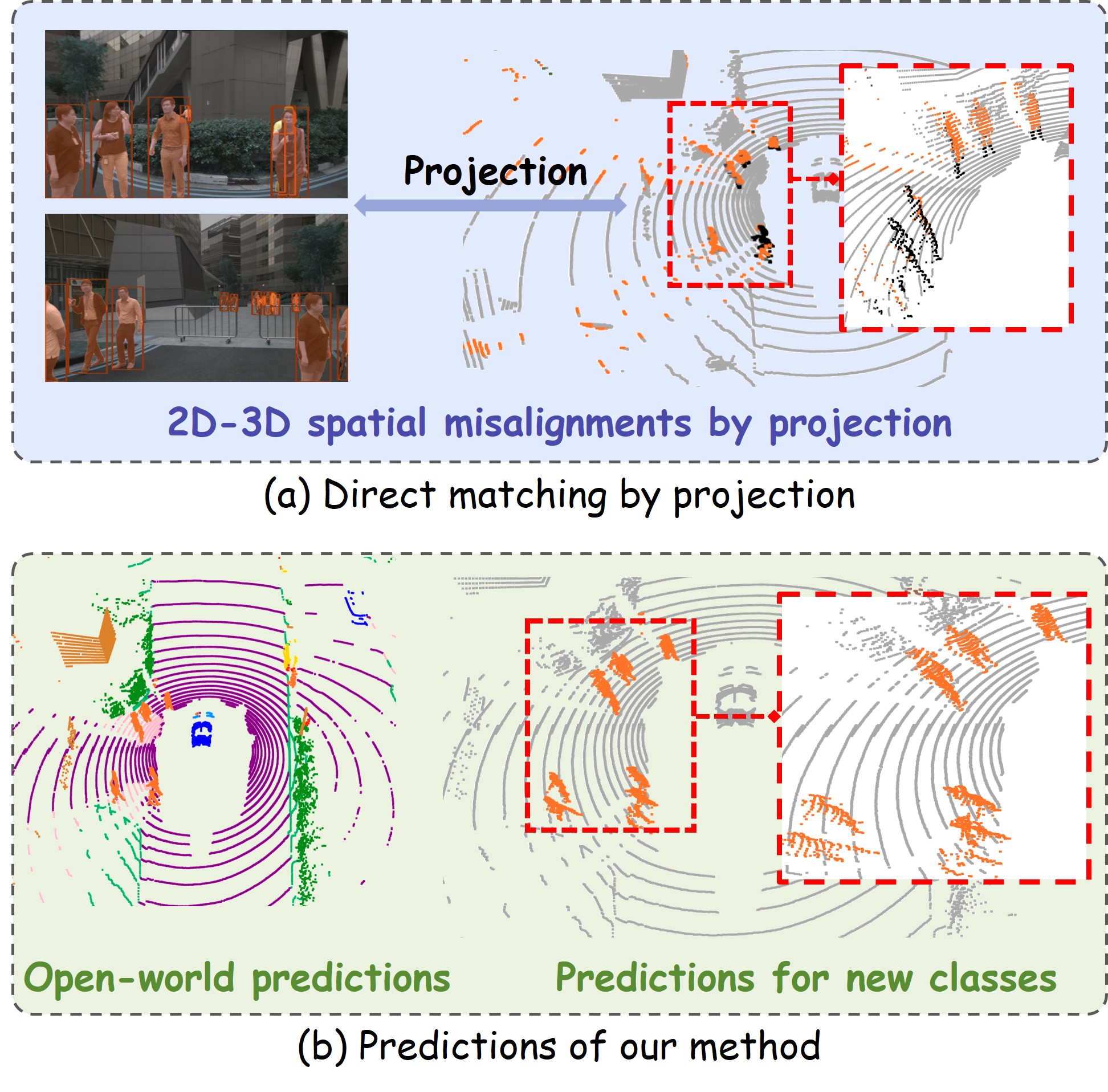}
	\caption{3D semantics directly projected from 2D semantics suffer from significant noise. Without any 3D annotations, our method achieves strong performance in 3D open-world semantic understanding by jointly training and co-embedding 3D points with pre-aligned images and text.}
	\label{fig:intro}
    \vspace{-1em}
\end{figure}

Recent breakthroughs in Visual Language Models (VLM) have ushered in new possibilities for bridging vision and language modalities, paving the way for open-world semantic understanding across domains. Extensions of VLM to 3D tasks, such as frozen-VLM labeling approaches \cite{zeng2023clip2, peng2023openscene}, utilize pre-trained vision-language alignment models to transfer open-vocabulary semantics to 3D data, enabling open-world scene understanding without manually labeled 3D annotations. However, these methods tend to neglect region-level feature distillation and often rely on resource-heavy processes, including extensive feature extraction, hierarchical fusion, and large-volume storage \cite{zhang2023clip}. Such inefficiencies limit their scalability to more advanced 3D networks and large-scale 3D scenes. On the other hand, pair-construction-based methods \cite{ding2024lowis3d, ding2023pla, regionplc} attempt to align point clouds and text representations, much like VLM training, by generating point cloud-text pairs for contrastive learning. While methods like RegionPLC \cite{regionplc} achieve state-of-the-art performance, creating point cloud-text pairs is a labor-intensive process that lacks the scalability of large-scale text-image datasets easily obtainable from the Internet. This fundamental bottleneck leaves pair-construction-based methods unable to fully capitalize on the open-world generalization capabilities of VLMs.

Nonetheless, point clouds frequently appear alongside image modalities with spatial and temporal alignment, suggesting an alternative strategy: use 2D models for image semantic labeling, followed by projection into 3D space. This approach introduces potential errors due to calibration and motion artifacts, resulting in noise and omissions in semantic data, as shown in Fig. \ref{fig:intro} (a). These challenges demand efficient methods to align text embeddings with point cloud features, minimizing projection inaccuracies. Given the spatial pairing of points and images, and the pre-existing image-text feature alignments in VLMs, images can act as bridges in a unified framework, easing the alignment of text with 3D point clouds.

Several critical challenges must be tackled to realize an open-world framework for 3D scene understanding with limited labeled 3D data: (i) co-embedding point cloud features with pre-aligned images and text into a shared feature space despite data constraints; (ii) addressing alignment errors caused by inaccurate projections between point clouds and images; and (iii) ensuring stable and effective multimodal training for robust performance across diverse scenes and categories.

To address these challenges, we propose \textbf{UniPLV}, a unified multimodal framework for open-world 3D scene understanding that eliminates the need for point cloud-text pair construction. UniPLV leverages pre-aligned visual language models and a point cloud understanding network, using images as a bridge to align point clouds and text in a shared feature space. To achieve this, we generate regional semantic labels for images using 2D foundation models and introduce logit and feature distillation techniques to align point cloud and image features. Additionally, we design a vision-point matching (VPM) module to refine point-pixel alignment and address projection errors. To tackle challenges such as gradient biases and long-tail distributions in multimodal training, we employ a two-stage training strategy with four task-specific losses: image-text alignment, point-text alignment, pixel-point matching, and distillation losses. Training begins with the image branch and transitions to joint training with balanced loss weights between modalities.
During inference, UniPLV requires only text descriptions and raw point clouds to achieve robust 3D semantic understanding. Extensive experiments demonstrate UniPLV’s state-of-the-art performance on open-world semantic segmentation tasks using nuScenes and ScanNet, with further evaluations conducted on Waymo and SemanticKITTI.

In summary, our contributions are as follows:
\begin{itemize}
\item We propose a unified multimodal contrastive learning framework for open-world 3D scene understanding without requiring point cloud-text pair construction.
\item We develop the vision-point matching module, logit distillation, and feature distillation to effectively co-embed point clouds, images, and text into a shared feature space.
\item We introduce a two-stage optimization strategy and four task-specific weighted losses to ensure stable and efficient multimodal training.
\item Our framework achieves state-of-the-art performance on multiple open-world benchmarks, including nuScenes, ScanNet, Waymo, and SemanticKITTI.
\end{itemize}

\section{Related Works}
\label{sec:related}


\textbf{Open-Vocabulary 2D Scene Understanding.} The recent advances of large visual language models have expanded the ability to understand 2D open-world scenes. There are two main directions: CLIP-based and Grounding.  CLIP-based methods usually replace the linear projection feature with the CLIP feature and  use CLIP for feature alignment, such as GLEE\cite{wu2024general}, DetCLIP series\cite{yao2022detclip, yao2023detclipv2, yao2024detclipv3}, RegionCLIP\cite{zhong2022regionclip}, OWL-ViT\cite{minderer2024scaling}.  The input of the Grounding tasks is a picture and a corresponding description. Grounding tasks output object boxes at the corresponding positions in the image based on different descriptions \cite{li2022grounded, zhang2022glipv2,liu2023grounding, ren2024grounding}. 

\noindent\textbf{Open-Vocabulary 3D Scene Understanding.} Open-world 3D scene understanding aims to recognize objects that are not annotated. With the success of visual-language (VL) models, such as CLIP \cite{radford2021learning}, numerous efforts \cite{chen2023clip2scene,takmaz2023openmask3d,zhang2023clip} have emerged to transfer the VL knowledge to 3D scene understanding. Clip2Scene \cite{chen2023clip2scene} utilizes frozen CLIP to obtain semantic labels of images, which are subsequently projected to guide a point cloud semantic segmentation. Further, Chen \textit{et al.} \cite{chen2023towards} employ CLIP to pseudo-label 2D image pixels and transfer those labels to 3D points, using SAM to reduce noise. Guo \textit{et al.} \cite{guo2024semantic} use a frozen 2D network to map to 3D points obtained through Gaussian Splatting. OpenMask3D \cite{takmaz2023openmask3d} employs a 3D instance segmentation network to create 3D masks, which are projected to obtain 2D masks. These 2D masks are then fed into CLIP to extract visual features and match them with textual features, ultimately obtaining 3D semantics. Since CLIP is trained on full-image and text alignments, its capability to comprehend specific regions is limited.  OpenScene \cite{peng2023openscene} facilitates point cloud and text alignment by projecting the prediction results from a frozen 2D vision model and employing distillation between image and point cloud features. However, OpenScene requires resource-intensive feature extraction and fusion,  and the image backbone is fixed during training, making it unable to be easily expanded to more advanced 3D networks and 3D scenes.  RegionPLC \cite{yang2024regionplc} and PLA \cite{ding2023pla}  achieve open world 3D scene understanding by constructing a large number of point clouds text pairs to train point clouds and text alignment.  
\begin{figure*}
	\centering
	\includegraphics[width=1\textwidth]{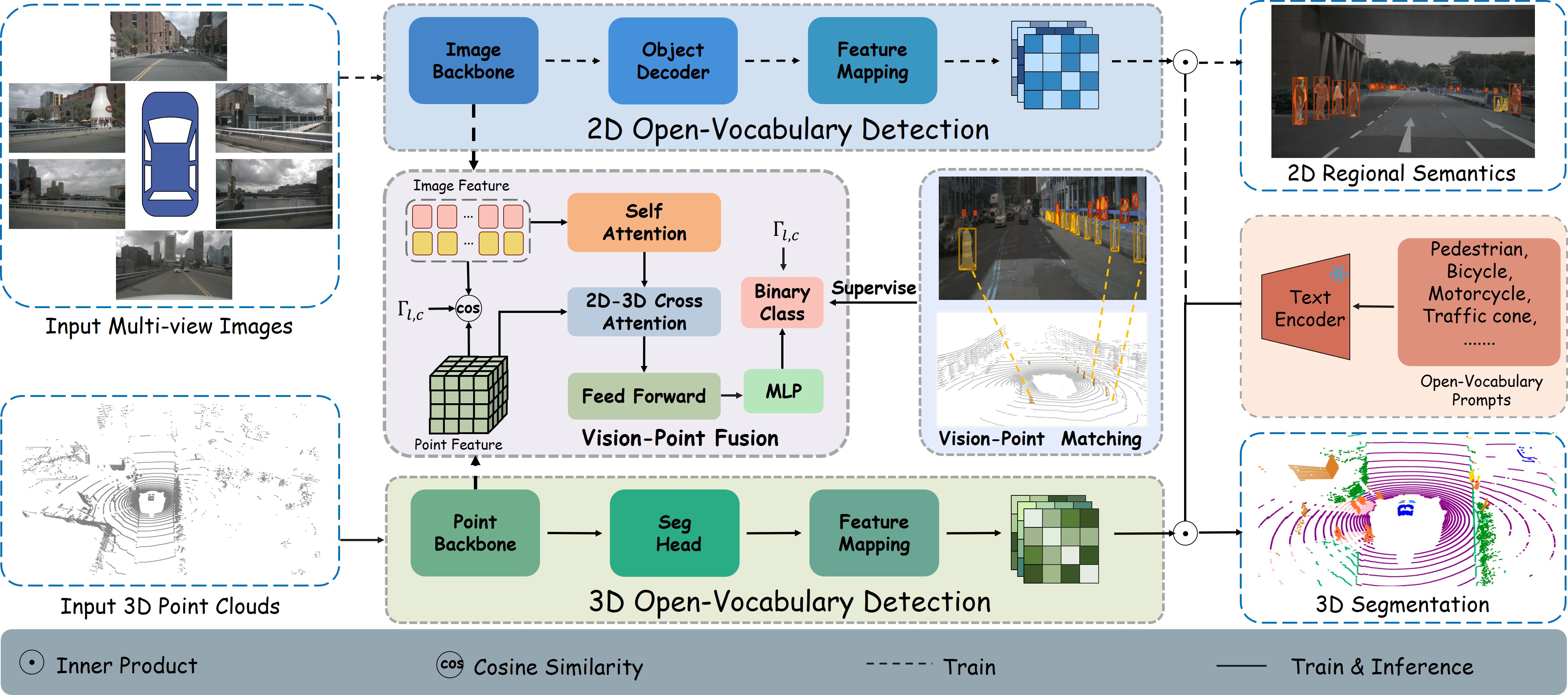}
	\caption{We introduce UniPLV, a multimodal learning framework that uses 2D foundation models to generate semantic labels for images as new class supervision. Category names are encoded as text prompts to create embeddings. For aligned point clouds and multi-view images, 2D and 3D branches extract features that are matched with text embeddings to categorize pixels and points. We distill image features to the point cloud to unify modalities and use a matching module to predict point-pixel correspondence. With just text and point clouds, Uni-PLV achieves effective 3D semantic understanding.}
	\label{fig:overview}
	\hfill
	\vspace{-1em}
\end{figure*}

\section{Methodology}

\subsection{Overview}
We present UniPLV, a new approach for open-vocabulary 3D scene understanding. UniPLV can identify novel categories without manual annotation, while also preserving the performance of base categories annotated manually. To be formalized, given a dataset of 3D point clouds containing categories $(\mathcal{C}_{b}\cup\mathcal{C}_{n})$, $\mathrm{s.t.~}\mathcal{C}_{b}\cap\mathcal{C}_n=\emptyset$. The method only accesses manual annotation of base categories $\mathcal{C}_{b}$ during training, but it can provide labels for both the base categories $\mathcal{C}_{b}$ and the novel categories $\mathcal{C}_{n}$ during inference. 

In contrast to previous works that achieved open vocabulary understanding by constructing 3D points-text pairs,  our work transfers the open vocabulary capability from 2D to 3D by utilizing image regional semantic labels from 2D foundation models without any additional data processing. With the 2D-3D spatial mapping and pre-aligned images and texts, we design a multimodal unified training framework that uses images as a bridge to embed point cloud features into the shared feature space of images and text. We introduce the main components of the proposed framework, the transformation of data flow, two knowledge distillation modules, and a vision-point matching module. Four task-specific optimization objectives are integrated in training. We introduce a multimodal and multi-task training strategy to ensure stable and efficient training of point cloud and image branches. During inference, this framework only requires point clouds and class descriptions as input to compute feature similarity, selecting the most similar class as the semantic prediction for each point.

\subsection{Image Data Processing}\label{subsection: data construct}
$\textbf{ Analysis. }$  Due to the more accessible and larger image-text pairing datasets, recent 2D open-world methods have made significant progress by aligning images and text through contrastive learning. Despite the scarcity of point clouds and the complexity of constructing point clouds-text pairs, point clouds and images are usually captured together and spatially aligned through calibration, which inspires us to use images as a bridge to align point clouds and text into a unified feature space. Considering that image-level alignment data generated by large-scale visual language models such as CLIP could not satisfy the needs of point-level scene understanding, we utilize 2D foundation models to obtain pixel-level correspondences between text semantics and image regions. We would like to emphasize that, despite the predicted image region semantics being noisy and presenting numerous misalignments when transferred to the 3D scene through projection, as illustrated in Fig. \ref{fig:intro}, our designed multimodal unified training framework, employing images as a bridge, can accurately identify and locate point clouds for given categories. 

\noindent$\textbf{ Region-Text Generation. }$  Intuitively, we extract instance and pixel semantics from 2D vision-language foundation models.  Specifically, given a set of images $\mathbf{I}=\{\mathbf{I}_{n}|n=1,...,N\}$ and the category list $\{t_c\}_{c=1}^C$ as input, where $t_c$ represents for the k-th category name, e.g., T = [``car", ``pedestrain", ``bicycle", ..., ``motorcycle"], the data pipeline output the bounding boxes $\mathbf{\hat{b}}$, instance mask $\mathbf{\hat{m}}$ and the category name $\mathbf{\hat{t}}$ for each image.   We use GLEE \cite{wu2024general} for instance mask and bounding box generation, which has been trained on large-scale datasets with high accuracy and generalization. We combine Grounding DINO \cite{liu2023grounding} and SAM2 \cite{ravi2024sam} to generate another set of instance labels.  Bounding boxes were generated via the Grounding DINO, and each box was further segmented using SAM2 to produce instance masks. We obtain the region-pixel-text pairs $\langle\mathbf{\hat{b}}, \mathbf{\hat{m}}, \mathbf{\hat{t}}\rangle$, as well as the point clouds spatio-temporally aligned to the images, to train the designed 3D scene understanding network transferred from regional image-text alignment. 2D semantic labels for the experimental results in this paper are from GLEE, and related experiments on Grounding DINO with SAM2 can be found in the supporting materials.

\subsection{MultiModal  Alignment}\label{subsection: Model achitecture}
\subsubsection{Model Framework}\label{subsubsection: Model composition}
The proposed UniPLV consists of a frozen text encoder,  image encoder-decoder, and point cloud segmentation branch, as illustrated in Fig. \ref{fig:overview}.  We input all category names as text prompts into the text encoder. We follow GLEE to apply global average pooling along the sequence dimension to get text embeddings $emb_t$. To support open-world understanding, we replace the classifier of the image decoder and the 3D segment head with the similarity measurement between perception features with text embeddings. Classification logits of 2D $L_{I}$ and 3D $L_{P}$ are computed as follows:
\begin{align}
	\label{eq:simi}
	L_{I}&=\phi(F_{I}(I)\cdot W_{i2t}, emb_t), \\
	\label{eq:simp}
	L_{P}&=\phi(F_{p}(P)\cdot W_{p2t}, emb_t),
\end{align}
where $\phi$ is the dot product, $\{W_{i2t},W_{p2t}\}$ are the mapping weights from image and point cloud to text feature dimension, $\{F_{I}, F_{p}\}$ are feature extractors for image branch and point cloud branch.

At each iteration $e$ of the training, we have a set of images $\{\mathbf{I}_{e}^k\in\mathbb{R}^{H\times W\times3} | k=1,...,K\}$ captured by $K$ cameras, where $H$ and $W$ are image dimensions. We also have the corresponding point cloud
$\mathbf{P}_e\in\mathbb{R}^{N_e\times3}$ with the  mapping matrix $\Gamma_{lc}$
from LiDAR coordinate to image coordinate.  Given the category list $\{t_c\}_{c=1}^C$, images $\{\mathbf{I}_e^k\}$ and the point cloud $\mathbf{P}_e$ as input, UniPLV could fine-tune the segmentation and detection of the image with the constructed regional image-text pairs $\langle\mathbf{\hat{b}}, \mathbf{\hat{m}}, \mathbf{\hat{t}}\rangle$, and provide the segmentation results of the point cloud corresponding to the given category list. The ultimate optimization goal of this framework is to embed  point cloud features and image-text features into a unified feature space through multimodal joint training, enabling the alignment of point cloud and text for open-world 3D scene understanding.

For the image and text branches, we load the second stage model of GLEE as pre-training weights to strengthen the alignment of text and images. During training, we fine-tune the image model using data constructed by 2D foundation models, exploiting the model's learning capability to eliminate noise from raw labels without additional processing. During iterative training, the model inherently engages in  feature clustering that identifies and learns the common attributes of the given category. This mechanism helps in filtering out noise introduced by false positives, thereby contributing to the effective cleaning of pseudo-labels.

\subsubsection{Vision-Point Knowledge Distillation}\label{subsubsection: Vision-Point Knowledge Distillation}
To construct the framework in which the image acts as a bridge for the joint embedding of point cloud features and pre-aligned image-text pairs into a unified feature space, we introduce two distillation modules: logit distillation and feature distillation. 

\noindent\textbf{Logit Distillation.}  The semantic classification logits $	L_{I}$ of image pixels are obtained through similarity measurement between the image feature and all given categories' txt embeddings as eq.\ref {eq:simi}. Similarly, the semantic classification logits $	L_{P}$ of point clouds are also obtained by calculating similarity with text as eq.\ref {eq:simp}. We design the logit distillation $	\mathcal{L}_{distill_{L}}$ supervising the point cloud classification logits $	L_{P}$  with novel class semantics $S_{C_{n}}$ predicted by the image branch as follows:
\begin{equation}
	S_{C_{n}} =\sigma(L_{I})\to Proj(\Gamma_{lc},\mathbf{P}), \\
\end{equation}
\begin{equation}
	\label{distill_L}  
	\begin{aligned}
		\mathcal{L}_{distill_{L}} =\mathcal{L}_{CE}(L_{P},S_{C_{n}}\cup Ig_{C_{b}})+ \\
		\mathcal{L}_{Dice}(L_{P},S_{C_{n}}\cup Ig_{C_{b}}), 
	\end{aligned}
\end{equation}
where $\sigma$ is the sigmoid function and $S_{C_n}$ are the category labels used to supervise the 3D segmentation of new classes, which obtained by mapping the predictions of the image through projection matrix $\Gamma_{lc}$; ${Ig}_{{C}_{b}}$ are masked labels of base class points; $\mathcal{L}_{CE}$ and $\mathcal{L}_{Dice}$ denote the  Cross-Entropy loss and Dice loss.

\noindent\textbf{Feature Distillation.} The alignment between images and text has been pretrained using large-scale data. To bridge the gap between point clouds and semantic text, we further distill the features of point clouds using image features. We exclusively distill 2D-3D paired points that simultaneously align in terms of spatial projection mapping and corresponding semantics. Feature distillation $\mathcal{L}_{distill_F}$ is performed based on similarity measurement as follows:
\begin{equation}  
	\label{distill_F}  
	\mathcal{L}_{distill_F} = 1- CoDist(F_{I}(I_m)\cdot W_{i2t}, F_{p}(P_m)\cdot W_{p2t})  \end{equation}
Where $CoDist(.)$ represents the cosine similarity function, $I_m$ and $P_m$ are 2D-3D pairings that align spatially and semantically.

\subsubsection{Vision-Point Matching Learning}\label{subsubsection: Vision-Point Matching Learning}
We introduce the vision-point matching (VPM) module to further learn fine-grained alignment between images and point clouds. It is a binary classification task that requires the model to predict whether the pixel-point pairs from the projection are positive (matched) or negative (unmatched). VPM mainly includes an attention encoder module and a binary classifier. Given the paring image features and the point cloud features, the image feature acts as the query vector $Q_I$, while the point cloud features serve as the key $K_P$ and value vectors $V_P$. Self-attention is applied to the image features, obtaining the image attention feature $Q_{attenI}$. Followed cross-attention between the  $Q_{attenI}$  and point cloud features as:
\begin{equation}\xi_i=\sigma(\frac{Q_{attenI}K_P}{\sqrt{d^\mathrm{\xi_i}}})V_P,\end{equation}
where $\xi_i$ is the obtained cross-attention features; $\sigma$ denotes softmax and $d^\mathrm{\xi}$ is the dimension of each $\xi$. The attention encoder can be formulated as:
\begin{equation}EN_{vpm}=\mathcal{FFN}(\mathrm{cat}(\mathrm{\xi}_1,...,\mathrm{\xi}_h)),\end{equation}
where  $\mathcal{FFN}$ is a feed-forward network, which consists of MLPs and activations. After the VPM encoder  gets the attention map $G_{attenIP}$ of the projection paring points and pixels, the matching probability is obtained by a binary classifier:
\begin{equation} \hat{f}_{map}=\operatorname{MLP}\left(G_{attenIP}\right),\quad f\in R^{r\times2},\end{equation}
where $r$ is the number of points and pixels that matched after projection; $\operatorname{MLP}$ denotes the multilayer perceptron for binary classification. The VPM module uses base and novel classes of pairing pixels and points for supervision during training.

\subsection{Optimization Objective}\label{subsection: Optimisation Objective}
We jointly train the alignment of image pixels and 3D points with text to achieve 3D open-world scene understanding. There are four task-specific  losses of our UniPLV: images-text alignment, points-text alignment, pixel-point matching, as well as  logits and feature distillation loss. The final overall loss is obtained by combining the above four types of losses with balancing weights as follows:
\begin{equation}
	\mathcal{L} =\beta \mathcal{L}_{image} + \delta  \mathcal{L}_{point} + \gamma \mathcal{L}_{distillation} + \gamma \mathcal{L}_{VPM}
\end{equation}
where $\beta,  \delta, \gamma$ denote the weights of losses; $ \mathcal{L}_{image}$ denotes the image-text alignment loss, which aims to predict boxes and semantic masks for images; $\mathcal{L}_{point}$ is the points-text alignment loss optimizing the segmentation of both base and new classes; $\mathcal{L}_{distillation}$ denotes the combination of logit distillation loss $\mathcal{L}_{distill_{L}}$ and feature distillation loss $\mathcal{L}_{distill_{F}}$; $\mathcal{L}_{VPM}$ represent the vision-point matching(VPM) loss to minimize projection errors between 3D points and 2D pixels. The VPM loss is then defined as:
\begin{equation}
	\begin{aligned}
		\mathcal{L}_{VPM} &= \textit{BCE}(\hat{f}_{map}, \mathbf{M}_{align}), \\
		\mathbf{M}_{align} &= 
		\begin{cases} 
			1 & \text{if } L_P==L_{I} \rightarrow Proj(\Gamma_{lc},\mathbf{P}),\\
			0 & \text{otherwise}.
		\end{cases}
	\end{aligned}
\end{equation}
where $\textit{BCE}$ denotes the Binary Cross-Entropy loss. The elements in $\hat{f}_{map}$, represent the matching labels between points and pixels, which are 1 if matched and 0 otherwise. The final loss is obtained by combining the above four types of losses with balancing weights. The specific formal expression of each loss can refer to the supporting materials.
\subsection{Multimodal Training and Inference}\label{subsection: Multi-Task Training}
We present a two-stage multi-task training strategy designed to train the multimodal framework UniPLV.

\noindent\textbf{Stage 1: Independent image branch training.} The training begins with a preliminary phase where we train the image branch independently, utilizing half of the total iteration steps to ensure a stable and robust foundation. We implement gradient clipping exclusively during the image branch training to prevent exploding gradients and stabilize the training dynamics.

\noindent\textbf{Stage 2: Unified multimodal training.} The second stage involves joint training of both the image and point cloud branches, employing distinct loss weights to balance their contributions effectively. Throughout the training process, we consistently utilize the AdamW optimizer, chosen for its adaptive learning capabilities and improved convergence properties. The optimizer parameters, specifically the learning rate and weight decay, are set differently for the image and point cloud branches depending on the backbone of each branch. This strategic differentiation in optimizer settings ensures that both branches are training according to their specific network structures and data characteristics, ultimately leading to superior performance in multimodal training tasks. The benefits of the two-stage training schemes, particularly for new classes, are detailed in the experimental section. 

The inference process is depicted in the Fig. \ref{fig:overview}. During inference, we can encode any arbitrary open-vocabulary categories as text queries and compute their similarity with the observed 3D points. Specifically, we associate each point with the category having the highest computed cosine similarity. Our method does not require processing images during inference, since we have distilled the image-text alignment to the point cloud.  

\section{Experiment}
\begin{table*}[htbp]
	
	\centering
	\caption{\textbf{Open-world 3D  semantic segmentation comparisons on nuScenes Benchmark}. Full supervision: Both base and new classes are gt labels. Open-world: Best open-world results are presented in bold.}
	\resizebox{\textwidth}{!}{
		\begin{tabular}{c|c|c|c|c|c|c|c}
			\toprule
			\multirow{2.6}{*}{Type}&\multirow{2.6}{*}{\centering Method} & \multicolumn{3}{c|}{B12/N3} & \multicolumn{3}{c}{B10/N5} \\ 
            \cmidrule{3-8}
			&&hIoU&mIoU$_B$& mIoU$_N$ & hIoU& mIoU$_B$ & mIoU$_N$ \\ 
            \midrule
			\multirow{3}{*}{\centering Full Supervision}& \multirow{1}{*}{SparseUnet32 \cite{graham20183d}} &75.2 & 76.6 & 73.8 &75.7& 76.6 & 74.8\\
			&\multirow{1}{*}{MinkUNet \cite{choy20194d}} &70.9& 74.6 &67.7 &72.3& 74.8 & 70.0 \\ 
			&\multirow{1}{*}{PTv3 \cite{wu2024point}} &77.2&79.9 &74.8 &77.6& 78.20 & 77.1 \\ 
            \midrule
			\multirow{8}{*}{\centering Open-World}
			&\multirow{1}{*}{3DGenZ \cite{michele2021generative}} &1.6& 53.3 &0.8 & 1.9&44.6 & 1 \\
			&\multirow{1}{*}{3DTZSL  \cite{cheraghian2020transductive}} &1.2& 21.0 & 0.6 &6.4&17.1 & 3.9 \\
			&\multirow{1}{*}{Ovseg-3D \cite{li2022language}} &0.6& 74.4 & 0.3 &0&71.5 & 0 \\
			&\multirow{1}{*}{PLA \cite{ding2023pla}} &47.7& 73.4 & 35.4 &24.3& 73.1 &14.5\\
			&\multirow{1}{*}{RegionPLC \cite{yang2024regionplc} (SparseUnet32)} &64.4& 75.8 & 56.0 &49.0& 75.8 & 36.3 \\
            \cmidrule{2-8}
			&\multirow{1}{*}{UniPLV(SparseUnet32)} &68.3& 75.9 &62.2 &65.3& 75.9 & 57.3 \\
			&\multirow{1}{*}{UniPLV(MinkUNet)} &64.1& 74.3 & 56.3 &64.5& 74.2 & 57.0 \\ 
			&\multirow{1}{*}{UniPLV(PTv3)} & \textbf{71.3}& \textbf{ 76.9} &  \textbf{66.5}& \textbf{69.6}&  \textbf{76.3} &  \textbf{64.0} \\ 
            \bottomrule
	\end{tabular}}

	\label{tab: comparsion nuscenes}
	
\end{table*}
\begin{table*}[htbp]
	
	\centering
	\caption{\textbf{Open-world 3D semantic segmentation comparisons on ScanNet}. Full Supervision: Both base and new classes are gt labels. Open-World: Best open world results are presented in bold.}
	\resizebox{\textwidth}{!}{
		\begin{tabular}{c|c|c|c|c|c|c|c}
			\toprule
			\multirow{2.6}{*}{Type}&\multirow{2.6}{*}{\centering Method} & \multicolumn{2}{c|}{B15/N4} & \multicolumn{2}{c|}{B12/N7} & \multicolumn{2}{c}{B10/N9} \\ 
            \cmidrule{3-8}
			&&mIoU$_B$& mIoU$_N$ & mIoU$_B$ & mIoU$_N$ & mIoU$_B$ & mIoU$_N$ \\ 
            \midrule
			\multirow{1}{*}{\centering Full Supervision}& \multicolumn{1}{l|}{SparseUnet32 \cite{graham20183d}} & 68.4 & 79.6 &  71.2& 71.6 & 76.5 & 71.9 \\ 
            \midrule
			\multirow{6}{*}{\centering Open-World}
			& \multicolumn{1}{l|}{3DGenZ \cite{michele2021generative}} & 56.0 &12.6 &35.5 & 13.3 &63.6 &6.6 \\
			&\multicolumn{1}{l|}{3DTZSL  \cite{cheraghian2020transductive}} & 36.7 & 6.1 &36.6 & 2.0 & 55.5 &4.2  \\
			&\multicolumn{1}{l|}{Ovseg-3D \cite{li2022language}} & 64.4 & 0.0&55.7 & 0.1 &  68.4 & 0.9 \\
			&\multicolumn{1}{l|}{PLA \cite{ding2023pla}} & 68.3& 62.4 & 69.5 &45.9 & 76.2 & 40.8\\
			&\multicolumn{1}{l|}{RegionPLC \cite{yang2024regionplc} (SparseUnet32)}& 68.2 & 70.7 & 69.9 & 66.6 & 76.3 & 55.6 \\
            \cmidrule{2-8}
			&\multicolumn{1}{l|}{UniPLV(SparseUnet32)} &  \textbf{68.4} & \textbf{75.7} &  \textbf{69.9} & \textbf{69.8} &  \textbf{76.6} &  \textbf{67.2} \\  \bottomrule
	\end{tabular}}
	
	\label{tab: comparsion scannet}
    \vspace{-1em}
	
\end{table*}
\subsection{Basic Setups}
\label{sec:Basic Setups}
\noindent$\textbf{ Datasets. }$ We evaluate our approach on three competitive outdoor datasets: nuScenes \cite{caesar2020nuscenes}, SemanticKITTI \cite{behley2019semantickitti}, and Waymo \cite{sun2020scalability} as well as the indoor dataset ScanNet \cite{dai2017scannet}. Following RegionPLC \cite{yang2024regionplc},  we evaluate the open-world understanding capabilities on two tasks: base-annotated open world (i.e., part of categories annotated) and annotation-free open world (i.e., no category annotated). For the base-annotated open-world task, we split the categories of the mentioned three datasets into base and novel classes, respectively.  As for nuScenes,  we follow the category partition of RegionPLC, disregarding the ambiguous category "other\_flat" and randomly dividing the remaining 15 classes into B12/N3 (i.e., 12 base and 3 novel categories) and B10/N5. We randomly split the category into B15/N2 for Waymo and B17/N2 for SemanticKITTI. Regarding the ScanNet dataset, we adopt the category partition of  PLA \cite{ding2023pla}.

\begin{table}[htbp]
	\centering
    \caption{Open-world evaluation on the SemanticKITTI and Waymo datasets.}
    \scalebox{1.14}{
	\begin{tabular}{c|c|c|c}
		\toprule
		Dataset &Method & mIoU$_B$$\uparrow$& mIoU$_N$$\uparrow$ \\ 
        \midrule
		\multirow{2}{*}{\centering Waymo }&Full Supervision& 74.3&81.8                  \\ 
        \cmidrule{2-4}
		&UniPLV& 74.0  &75.4                  \\ 
        \midrule
		\multirow{2}{*}{\centering SemanticKITTI }& Full Supervision                                           & 57.7                      & 77.4                  \\
        \cmidrule{2-4}
		&UniPLV&  57.4  &62.3  \\ 
        \bottomrule
	\end{tabular}
    }
	\vspace{-1em}
	\label{tab: WS eval}	
\end{table}

\noindent$\textbf{ Implementation Details and Evaluation Metrics. }$ We build the text and image branch upon GLEE \cite{wu2024general},  adopting CLIP \cite{radford2021learning} text encoder to process text descriptions and MaskDINO \cite{li2023mask} as  image encoder-decoder. We utilize the MinkUNet \cite{choy20194d}, SparseUnet32 \cite{graham20183d}, and PTv3 \cite{wu2024point} as point cloud backbone. We use the ADAM optimizer with a batch size of 32 on 16 NVIDIA A800 GPUs. We use the mean intersection-over-union (mIoU) and harmonic mean IoU (hIoU) as the evaluation indicators. 

\subsection{Quantitative Results}

\subsubsection{Base-Annotated Open-World}
Table \ref{tab: comparsion nuscenes} shows major comparison methods, among which RegionPLC is the current SOTA method. Since most public results are reported on nuScenes, the primary comparison focuses on this benchmark. For SemanticKITTI and Waymo, we present the gap with full supervision and visual analysis. 

\noindent\textbf{nuScenes Results.} For a fair and detailed comparison, we present the results of three point cloud segmentation backbones in Table \ref{tab: comparsion nuscenes}. Using the same point cloud backbone, SparseUnet32 as RegionPLC, our method largely lifts the mIoU of unseen categories by 6.2\% $\sim$21.8\% among various partitions on nuScenes without constructing any point-text pairs. When utilizing the higher-performance backbone network, PTv3, our approach achieves a mIoU gain of 10.5\% $\sim$27.7\% for the new classes compared to RegionPLC. Even with a lower performance backbone, MinkUNet, our method still outperforms RegionPLC by 0.3\% $\sim$20.7\% for new classes. While achieving significant improvements in the new classes, we can keep the mIOU of the base classes fluctuating within 0.3\% $\sim$3\% compared to fully supervised performance. Comparing the 3DGenZ and 3DTZSL, which omit language alignment, our method even obtains 60.1\% $\sim$65.9\% gain on mIoU of new classes. The results for different backbone substitutability demonstrate the adaptability and integrability of our method.

\begin{table}[htbp]
	\centering
    \caption{Comparisons for annotation-free task on nuScenes.}
    \scalebox{1.2}{
	\begin{tabular}{c|c|c}
		\toprule
		\multicolumn{1}{c|}{Dataset} & Method & mIoU$\uparrow$ \\ 
        \midrule
		\multirow{5}{*}{nuScenes}& \multicolumn{1}{l|}{CLIP2Scene} & 20.8 \\ 
		~& \multicolumn{1}{l|}{OpenScenes} & 42.1 \\
		~&\multicolumn{1}{l|}{MSeg Voting} & 31.0 \\
		~&\multicolumn{1}{l|}{2D-3D projection} & 28.6 \\ 
        \cmidrule{2-3}
		~&\multicolumn{1}{l|}{UniPLV} & \textbf{56.9} \\ 
        
        \bottomrule
	\end{tabular}
    }
	\label{tab:anno}
	\vspace{-1em}
\end{table}

\noindent\textbf{ScanNet Results.}
As shown in Table \ref{tab: comparsion scannet}, UniPLV outperforms the previous state-of-the-art method, RegionPLC. Specifically, in the performance on unseen categories, our approach exceeds RegionPLC by a margin of 3.2\% to 11.6\%, while the gap relative to fully supervised methods is only 1.8\% to 4.7\%.

\noindent\textbf{SemanticKITTI and Waymo Results.}  As shown in Tab \ref{tab: WS eval}, partition B15/N2 on Waymo shows only a 6.4\% difference in new classes performance compared to the full supervision, which demonstrates the robustness and practicality of our approach in open-world 3D  scene understanding. On the SemanticKITTI dataset, the performance of the B17/N2 partition differs from fully supervised classes by 15.1\%. It should be pointed out that the SemanticKITTI dataset only has forward-looking images, and we only capture point clouds at a forward angle of $120^\circ$ for experimentation, which leads to a large performance gap with full supervision. The results on these two datasets demonstrate the broad applicability of our proposed method in different scenarios.

\begin{figure*}[h]
	
	\centering
   
	\includegraphics[width=1\textwidth]{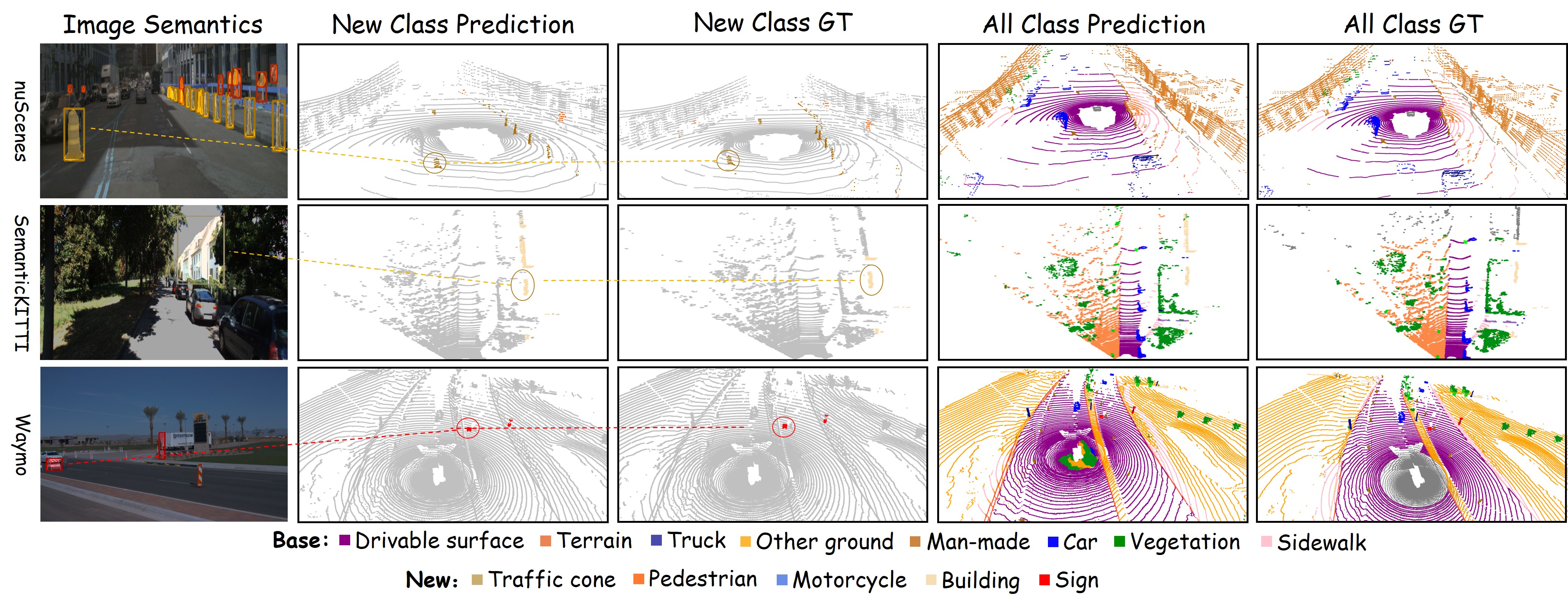}
	   \caption{Qualitative results of the base-annotated task.}
	
	\label{fig:pre}
	
\end{figure*}
\begin{table*}
	
	\centering
    \caption{Optimization analysis: including ablation study for two-stage optimization strategies and various loss functions.}
	\scalebox{1.3}{
	\begin{tabular}{cccccc|c|c}
		\toprule
		One-Stage & Two-Stage & $\mathcal{L}_{logits}$ & 	$\mathcal{L}_{distill_N}$  & $\mathcal{L}_{distill_F}$ &  $\mathcal{L}_{VPM} $  &  mIoU$_B$$\uparrow$  &     mIoU$_N$$\uparrow$ \\ 
        \midrule
		\checkmark   &                      &\checkmark     &                      &                     &                         &        73.2                   &        57.3      \\ 
		\checkmark   &                      &\checkmark     &\checkmark   &\checkmark   &\checkmark     &        74.1                   &        58.2      \\
		&\checkmark   &\checkmark     &                      &                      &                        &        76.6                  &        60.1      \\
		&\checkmark   &\checkmark     &\checkmark   &                      &                        &        76.4                  &        62.6       \\
		&\checkmark   &\checkmark     &\checkmark   &\checkmark   &                        &        76.8                &        65.2      \\
		&\checkmark   &\checkmark     &\checkmark   &\checkmark   &\checkmark     &        76.9                 &        66.5       \\
        \bottomrule
	\end{tabular}
    }
	
	\label{tab:Optimization analysis}
	\vspace{-1em}
\end{table*}

\subsubsection{Annotation-Free Open-World Segmentation.}

We evaluate the annotation-free capability of our method on the nuScenes validation set, which consists of 16 categories. We freeze the image branch of the proposed UniPLV and load the pre-trained weights from GLEE. Without any 2D or 3D annotations, we have successfully transferred 2D open-world capabilities to 3D through distillation and matching learning.  As shown in Table \ref{tab:anno}, we compare our approach to the most closely related work on zero-shot 3D semantic segmentation: CLIP2Scene \cite{chen2023clip2scene}, OpenScene \cite{peng2023openscene}, and MSeg \cite{lambert2020mseg} Voting. UniPLV achieves better performance than the above three methods. UniPLV outperforms CLIP2Scene with 36.1\%, OpenScene with 14.8\%, and MSeg Voting with 25.9\%, respectively.

\subsection{Qualitative analysis}
We further demonstrate the open-world 3D understanding capabilities of the proposed method through visualization. As shown in Fig. \ref{fig:pre}, our method successfully identifies and locates new categories while ensuring the recognition accuracy of base categories. The object detected by the image can be located by the point cloud network, and some false detections in the image can also be adaptively filtered out by the point cloud network. For some relatively small and distant targets with few points, such as traffic cones and distant pedestrians, our method still locates and identifies them.

\subsection{Ablation Studies}
\label{abstudy}

The ablation experiments are all conducted on nuScenes based on the category partition B12/N3.
\noindent$\textbf{ Multimodal Learning Framework. }$ To verify the contrastive learning ability of the proposed framework, we directly project point clouds into images to get 2D pixel semantics predicted by 2D foundation models. As shown in Table \ref{tab:anno}, our proposed framework outperforms the projection-only approach by 37.9\% of mIoU on new classes. This experiment shows that the proposed framework can align the three modes well with limited point cloud data.

\noindent$\textbf{Alignment and Optimization.}$ Table \ref{tab:Optimization analysis} presents the impact of different modules, corresponding optimization losses, and the two-stage training procedure. In comparative experiments between one-stage and two-stage training, two-stage training yields an improvement of 2.8\% $\ sim$8.3\% for both base and new classes.  Based on the loss $\mathcal{L}_{logits}$, adding an independent distillation $\mathcal{L}_{distill_N}$ for the new class resulted in a 2.5\% increase. The feature distillation loss $\mathcal{L}_{distill_F}$  from pixels to points results in a 2.6\% improvement. The vision-point matching learning loss $\mathcal{L}_{VPM} $ brings 1.3\% mIoU gains in the new classes. The experimental results demonstrate the effectiveness of our designed 2D to 3D knowledge transfer alignment strategy and provide a valuable reference for multimodal fusion tasks involving point clouds and images.

\section{Conclusion and Future Work}

We present a unified multimodal learning framework, termed UniPLV, for open-world 3D scene understanding that does not require point cloud text pairs. Leveraging image as a bridge, we propose logits distillation, feature distillation, and modal matching modules. We further introduce four task-specific losses and a two-stage training procedure for stably multimodal learning. Our method significantly outperforms SOTA methods on nuScenes and ScanNet. Besides, experimental results on different 3D backbones as well as on Waymo and SemanticKITTI datasets demonstrate the scalability and lightweight of our method. We will improve and quantify the image branch in the future, so that the proposed framework can achieve both 2D and 3D open-world scene understanding tasks .  The point cloud branch can also be replaced by an occupancy prediction network.


\bibliographystyle{IEEEtran}
\bibliography{root.bib}

\end{document}